\documentclass[letterpaper]{article} 

\usepackage{aaai2026}  
\nocopyright

\usepackage{times}  
\usepackage{helvet}  
\usepackage{courier}  
\usepackage[hyphens]{url}  
\usepackage{graphicx} 
\usepackage{subcaption}

\usepackage{xcolor}

\usepackage{natbib}  
\usepackage{caption} 
\usepackage{url}

\usepackage{caption}
\usepackage{booktabs}
\usepackage{tabularx}
\usepackage{array}
\newcolumntype{L}[1]{>{\raggedright\arraybackslash}p{#1}}
\usepackage{subcaption}
\usepackage{amsmath}
\usepackage{amssymb}
\usepackage{breqn}
\usepackage{fontspec}

\newfontfamily\hindifont[
  Script=Devanagari,
  Path=./,
  Extension=.ttf,
  AutoFakeBold=2
]{NotoSansDevanagari-Regular}

\newcommand{\texthindi}[1]{{\hindifont #1}}

\usepackage{multirow}
\usepackage{amsthm}

\usepackage{algorithm}
\usepackage{algorithmic}

\usepackage{fontspec}
\usepackage{polyglossia}
\setmainlanguage{english}

\usepackage{multirow}

\usepackage{changepage}

\usepackage{newfloat}
\usepackage{listings}
\DeclareCaptionStyle{ruled}{labelfont=normalfont,labelsep=colon,strut=off} 
\floatstyle{ruled}
\newfloat{listing}{tb}{lst}{}
\floatname{listing}{Listing}
\title{\textit{You Are What You Read}: \\Indian Political News Media Bias Dataset for Hindi Language}
\author{
    Kushal Trivedi\textsuperscript{\rm 1}, 
    Mayank
    Mahor\textsuperscript{\rm 1}
    Dhruv Sanghvi\textsuperscript{\rm 1},
    Roshni Chakraborty\textsuperscript{\rm 1}
}
\affiliations{
    \textsuperscript{\rm 1}Atal Bihari Vajpayee - Indian Institute of Information Technology, Gwalior, India \\
    kushal.trivedi.2110@gmail.com
}

\begin{document}

\maketitle

\begin{abstract}
Media bias understanding and detection remains an under-researched problem for Indian context, specifically for Indic languages, such as, Hindi, which is widely spoken in India. In this paper, we introduce a comprehensive, multi-annotator, explainable dataset consisting of approximately $448$ unique articles from four leading Indian news agencies publishing in Hindi. We focus on Hindi, the most spoken and an official language of India. The study of media bias is particularly relevant in the Indian context due to the country’s linguistic diversity and large population, where the same event is often reported by multiple news agencies in different languages, thereby introducing scope for bias. The dataset comprises Hindi political news articles annotated by $14$ experts and classified into four bias categories  - selection bias, framing bias, clickbait bias and omission bias - when an article is identified as biased or unbiased. For each biased article, annotators also provide a textual explanation, enabling future semi-automated approaches to incorporate explainability when language models reason over the dataset. Cohen’s Kappa scores were $0.81$ for simple bias/no-bias classification reflecting strong inter-annotator agreement. For the multi-label task, the Cohen’s Kappa values are $0.89$, $0.75$, $0.81$, and $0.79$ for clickbait, selection, framing, and omission biases, respectively, resulting in an average Kappa of $0.80$, which also reflects strong agreement.The dataset and replication code are publicly available at \textbf{\url{https://github.com/KushalTrivedi19032005/Indic-Media-Bias-Detection-Dataset}}.


\end{abstract}

\section{Introduction}
\par News, whether in physical or electronic form, remains the primary medium for communicating recent events to the public \cite{fan2025skeptik}. With increasing digitization, users now have continuous access to news from multiple news media agencies, which lead to high competition for user attention\cite{chakraborty2019tweet}. While audiences rely on news agencies for fair and high-quality reporting, there have been numerous instances of deviation from journalistic standards, including deliberate misinformation or omission of context, lack of fact-checking, clickbait headlines, biased framing, and politically aligned reporting \cite{kuuse2025crowdsourced, luusi2024political}. Media bias refers to the \textit{non-objective presentation of news through mechanisms such as selective reporting, omission of relevant facts, or slanted perspectives based on demographic dimensions including political ideology, gender, religion, geography, or ethnicity} \cite{spinde2023media}. Alternatively, \cite{entman1993framing} defines media bias as \textit{when journalists systematically frame reality in a way that promotes a particular interpretation, evaluation, or solution to an issue}. Regardless of its form, media bias can adversely shape public opinion, facilitate misinformation, contribute to echo chambers, and intensify societal polarization \cite{arifah2025social}. Consequently, identifying, detecting, and mitigating media bias is essential to ensure an informed readership \cite{hamborg2019automated}. However, the extensive information overload makes it increasingly difficult for users to manually distinguish between biased and unbiased news, often leading to skewed news consumption \cite{rodrigo2024systematic}. Automated bias detection is highly challenging as it requires identification and understanding of the underlying implicit attributes and its various forms, such as framing bias, selection bias, and presentation bias. Each type requires specialized bias detection approaches. This further aggravates for in a country as vast and linguistically diverse as India \cite{sen2022analysis}.



\par According to Lokniti CSDS - a network of political scientists conducting survey-based studies on Indian politics and democracy - around 82\% of independent journalists surveyed believe that Indian media favors a political party.\footnote{\url{https://maktoobmedia.com/india/lokniti-csds-study-80-journalists-believe-media-covers-modi-govt-too-favourably/}}. Further, the World Press Freedom Index 2025 ranks India 151st out of 180 countries,\footnote{\url{https://www.outlookindia.com/national/india-ranks-151-out-of-180-countries-in-world-press-freedom-index-2025-rsf-calls-it-one-of-worlds-most-dangerous-countries}} highlighting significant economic pressure on journalists and a high concentration of media ownership, both of which threaten independent and objective reporting. This concern is particularly relevant in today’s digital era, where a 2022 study by Oxford University Press (OUP) found that 54\% of Indian youth turn to social media when seeking factual information.\footnote{\url{https://www.livemint.com/news/india/despite-misinformation-concern-54-indians-derive-factual-information-from-social-media-reveals-oup-study-11656420664780.html}} Therefore, in order to build automated systems for media bias detection and mitigation for Indian context and specifically, in Indic languages, such as, Hindi, there is a requirement of specific extensive datasets. To quantify media bias in a systematic manner, the availability of comprehensive and reliable datasets is a fundamental prerequisite. While several datasets exist at both international and national levels for bias classification and related tasks in high-resource languages, the Indian context presents unique challenges. The high variance and richness of Indic language differences along with India's complex political landscape featuring a multi-party system unlike the largely binary political ideologies in other English speaking countries makes direct application of bias detection approaches ineffective. Furthermore, translation-based approaches fail to account for culturally and linguistically specific nuances of Indian society \cite{puro2026challenges} or AI-driven machine translation might lose sensitivity\cite{anik-etal-2025-preserving}.


\par Despite being India’s official language, Hindi remains a low-resource language in this domain \cite{sabane2023breaking}. Existing research on Indian news media bias largely focuses on English-language outlets, primarily analyzing Indian English newspapers \cite{sen2022analysis}. The high variance in Indian languages makes it difficult to build a system that can work for media bias irrespective of the Indian language. Therefore, given the unavailability of datasets and absence of research works in this low resource domain, such as, media bias detection and mitigation, we focus on one of the primary Indic languages, i.e., Hindi. Although recent studies have explored the use of large language models (LLMs) as alternatives to manual annotation \cite{horych2025promises, raza2024dataset}, manual annotation continues to outperform automated approaches in tasks requiring nuanced judgment, domain expertise, and ethical consideration \cite{garg2025portrait}. These models might lack robust capabilities for reasoning-intensive tasks, including bias classification and detection, for Indian context (as shown through an example in Table~\ref{tab:tab1}). With respect to the specific example as shown in Table~\ref{tab:tab1}, we observe the annotator labels are presented as the ground truth, whereas the language models incorrectly predict several bias types, such as clickbait, emotional, and ideological biases. Although identifying labeled data requires manual annotation, it is time and cost intensive. Therefore, it is particularly challenging to build automated models, especially train large language models, for media bias detection in Indian languages. Therefore, in this paper, we introduce the first Hindi media political bias benchmark dataset, covering four distinct types of bias. To the best of our knowledge, this is the first dataset in Indian context that covers an exhaustive types of biases. This dataset aims to support the research community in developing robust and trustworthy automated systems for monitoring and mitigating news media bias in Indian context.

\begin{table*}[!t]
\centering
\caption{Human vs LLM Classification Comparison}
\label{tab:tab1}
\footnotesize
\begin{tabular}{|p{3.8cm}|p{4.0cm}|p{3.8cm}|p{4.0cm}|}
\hline
\textit{Original Article (Hindi)} &
\textit{Translated Article (English)} &
\textit{Ground Truth} &
\textit{LLM's Prediction} \\
\hline

\parbox[t]{\linewidth}{\raggedright
\texthindi{\textbf{दल }}A\texthindi{\textbf{ ने निगेटिव नैरेटिव चलाया, जनता ने रिजेक्ट कर दिया; हरियाणा के नतीजों पर व्यक्ति }}A\texthindi{\textbf{ और क्या बोले?}}

\par
\texthindi{हरियाणा में दल }B\texthindi{ ने हैट्रिक लगाकर लगातार तीसरी बार सरकार बनाने का मेंडेट हासिल कर लिया है। राज्य के चुनावी नतीजों पर देश भर में राजनीतिक हलचल मची है। दल }B\texthindi{ ने तीसरी बार जीत हासिल करके ऐतिहासिक रिकॉर्ड कायम कर दिया। इस पर बिहार में सियासी बयानबाजी है।}

\par
\texthindi{दल }C\texthindi{ सुप्रीमो के व्यक्ति }C\texthindi{ नेता प्रतिपक्ष व्यक्ति }D\texthindi{ ने जहां चुनाव परिणाम पर हैरानी जताई तो दल }B\texthindi{ ने ढोल नगाड़े बजाकर जश्न मनाया।}

\par
\texthindi{इधर दल }C\texthindi{ के राष्ट्रीय अध्यक्ष एवं व्यक्ति }E\texthindi{ सरकार में पंचायती राज मंत्री व्यक्ति }F\texthindi{ ने दल }A\texthindi{ को बड़ी नसीहत देते हुए कहा कि जनता ने इस पार्टी को रिजेक्ट कर दिया है जबकि व्यक्ति }E\texthindi{ का डंका चारों ओर बज रहा है।}
}
&

\parbox[t]{\linewidth}{\raggedright
\textbf{Party A ran a negative narrative, the public rejected it; what else did Person A say on the Haryana results?}

\par
In Haryana, Party B has achieved a hat-trick and secured a mandate to form the government for the third consecutive time. The election results have triggered political stir across the country. By winning for the third time, Party B has created a historic record. Following this, political statements have intensified in Bihar.

\par
While Party C supremo Person C and Leader of the Opposition Person D expressed surprise over the election results, Party B celebrated by beating drums and rejoicing. Meanwhile, Party C president and Union Minister for Panchayati Raj in the Person E government, Person F, strongly advised Party A, saying that the public has rejected this party, whereas Person E's influence is resonating everywhere.
}
&

\parbox[t]{\linewidth}{\raggedright
\textbf{Selection Bias}

\textbf{Framing Bias}

\textbf{Stance Bias}

\textbf{Omission Bias}

\par\medskip
1. \texthindi{"इससे यह भी साबित होता है कि अब दल }A\texthindi{ ज्यादा नहीं चलेगी।"} \\(\textit{English Translation}: This also proves that Team A won't last long.)
$\Rightarrow$ Stance bias

\par
2. Omission bias $\Rightarrow$ Negative comments are not mentioned

\par
3. Framing bias $\Rightarrow$ Victory is foregrounded before political reactions

\par
4. Selection bias $\Rightarrow$ Maximum space is given to Person E, a Party B ally
}
&

\parbox[t]{\linewidth}{\raggedright
\textbf{Framing bias}:  
The article frames Party B's victory as a ``historic record'' and contrasts it with Party A's failure.

\par\medskip
\textbf{Linguistic bias}:  
Loaded phrases like \texthindi{"जनता ने रिजेक्ट कर दिया"}, \texthindi{"चारों ओर डंका बज रहा है"}, and \texthindi{"बड़ा जुबानी हमला"} are used.\\
(\textit{English Translation}: Loaded phrases like "The public has rejected it", "There is a lot of noise everywhere" and "A massive verbal attack are used.")

\par\medskip
\textbf{Ideology bias}:  
The narrative aligns with a pro-Party B perspective.

\par\medskip
\textbf{Stance bias}:  
Opposing voices are framed defensively, while Party B-aligned views are reinforced.

\par\medskip
\textbf{Emotional bias}:  
References to celebration, rejection, humiliation, and attacks are used to evoke pride, triumph, or resentment rather than just inform.

\par\medskip
\textbf{Clickbait bias}:  
The highlighted line \texthindi{“ये भी पढ़ें: व्यक्ति} E \texthindi{के व्यक्ति} G \texthindi{चावल भोज में बवाल, देखें} Video” is designed to provoke curiosity and drive clicks rather than add factual value. \\
(\textit{English Translation:} Chaos at Person G's rice feast, watch video!)

\par\medskip
\textbf{Omission bias}:  
The article does not deeply explore neutral electoral data, voter issues, or independent analysis, limiting alternative explanations for the results.
}
\\
\hline
\end{tabular}
\end{table*}


\par The rest of the paper is organized as follows, related works in Section \ref{s:rw}, dataset collection and annotation detail in Section \ref{s:data} followed by discussions in Section \ref{s:res}. We discuss a brief overview of FAIR principles in Section \ref{s:fair} and finally, conclusions in Section \ref{s:con}.

\section{Related Works} \label{s:rw}
\par Bias detection in news media is specifically relevant considering the adverse impact on public opinion shaping and society \cite{fan2019plain,spinde2023media}. Media bias based research works tackle different forms, such as, selection bias, information bias, editorial bias, framing bias, demographic bias, etc \cite{rodrigo2024systematic}.  Existing research works on framing based media bias detection  could be segregated into feature based understanding, such as, usage of kill verbs, lexical cues, sentiment analysis \cite{pryzant2020automatically,bhatia2018topic}, contextual representation of bias inducing words and sentence level understanding \cite{raza2024unlocking,lei2024sentence}. Recent works on transformer based bias detection integrate pretraining datasets through Distant Supervision Learning \cite{spinde2021mbicmediabias}, Multi-task Learning \cite{spinde2022exploiting}, domain adaptive dataset \cite{raza2024dataset}. While these approaches have shown high effectiveness for framing bias at article level, they do not discuss the interpretability of the model’s decision which can aid in bias mitigation and understanding \cite{noh2025biassist}. Additionally, these approaches do not provide detail of the rationales that can enhance the interpretability \cite{spinde2025enhancing}. Furthermore, these models are not directly applicable to Indian languages due to the requirement of understanding and interpretation of framing bias which can not be resolved by translation compiled with the existing research works. 

\par There are several existing research works that focus on bias detection and characterization in text with respect to demographic bias, such as, racial bias\cite{kambhatla2022surfacing,navon2025fairness}, gender bias \cite{cheng2023marked}, etc. While these approaches mainly focus on bias detection in user generated contents, such as, social media posts, hate speech, etc., these provide significant understanding of the characteristics of bias and possible ways of detection \cite{cheng2023marked}. For example, \cite{navon2025fairness} highlights the role of lexical features, such as, readability measures and Yule’s measure in automated gender bias detection and further, discusses human mediated mechanisms for bias mitigation. 

However, the high variance in Indian society requires specific handling of demographic bias \footnote{\url{https://socialwork.institute/social-science-concepts/india-demographic-diversity-facts-figures/#population-composition-and-trends}} \cite{10.1007/978-3-031-19097-1_33}. This could be with respect to the variation in population density across regions. For example, Delhi, Mumbai, Kolkata, and Chennai represent some of the most densely populated districts with over 10,000 people per square kilometer, while remote areas in Arunachal Pradesh and Ladakh have densities as low as 20 persons per square kilometer. This geographic diversity means social work approaches, such as the number of news articles for a particular group or a social issue, or bias detection and mitigation through automated approaches, must be preferably tailored to local contexts rather than applying uniform solutions across the country. Similarly, disparities in gender and age also create a requirement for such contextual adaptation.

\par Subsequently, news media agencies are required to filter out the particular information from all the available information to be reported. This selection might lead to pruning of various news events or select a subset of information given a news event. While this selection is a necessity for news publishing, the reasoning might be based on ideological leaning which can adversely impact the representation of information, demographic groups, political leaning, etc. This is referred to as selection bias or. information coverage bias. Existing research works compares the latent representation of a news event based on the news articles published with respect to it from different news media sources to the particular news event representation for a news media source \cite{zhou2025calibrated,wang2025media}, explore theoretical understanding of  journalism, communications, and political science, revealing key insights into usability and functionality, comparison of right leaning based political parties and  and explored narrative variation through stance and topical based analysis \cite{kuila2024news}. Subsequently, bias mitigation based approaches, such as, one of the earliest and most common approaches to bias mitigation is supervised fine-tuning using curated, balanced datasets. For example, in \cite{zhao2018gender, khan2025investigating}, Zhao et al. introduced Counterfactual Data Augmentation (CDA) for coreference resolution, where gender-swapped sentence pairs are used to reduce gender bias in model predictions. Similarly, crafting balanced and inclusive prompts during fine-tuning has been shown to improve model fairness and diversity in output.

\par Therefore, irrespective of the type of bias and context, the requirement of labelled datasets with rationale for development of automated trustworthy media bias detection and/or mitigation approaches. Existing media bias datasets include \textit{MBIB} -- Media Bias Identification Benchmark \cite{10.1145/3539618.3591882} is a large benchmarking dataset created by combining 15 existing datasets. It classifies statements, sentences, and tweets into nine bias categories, such as, linguistic bias, text-level context bias, reporting-level context bias, cognitive bias, hate speech, racial bias, fake news, gender bias, and social bias. Subsequently, \textit{MBIC} \cite{spinde2021mbicmediabias}, i.e., \textit{Media Bias Annotation Dataset Including Annotator Characteristics} consists of 1,700 statements representing various forms of media bias. Each statement was annotated by ten annotators and includes labels at both the word and sentence levels. To the best of our knowledge, MBIC is the first media bias dataset to provide detailed information about annotator characteristics and individual backgrounds. In addition, \cite{spinde-etal-2021-neural-media} proposed a neural approach to media bias detection using a multi-stage annotation process. Annotators first identified words or phrases that induced bias, then labeled sentences as biased or non-biased, and finally categorized them as factual, opinionated, or mixed. There has been very few works for other than English language, such as, \textit{Implicit Media Bias Dataset} \cite{LEE2023128219} comprised of 24,576 news articles covering non-political events which were annotated for objectivity, fairness, unbiasedness, and political orientation, and was classified as conservative, neutral, or progressive.



\par Substantial work has been carried out to create annotated datasets for media bias detection. However, most of this effort is concentrated on high-resource languages, with only a limited number of notable works addressing low-resource domain based work, such as, media bias detection and mitigation for Indian context. Additionally, it was highlighted political bias identification and mitigation is extremely challenging compared to other forms of bias, such as, hate speech, racial bias, and gender bias \cite{spinde2021mbicmediabias}.  Compared to international research on news media bias detection, work in the Indian context remains relatively limited. Creating annotated datasets for low-resource domains is both labor-intensive and costly. Existing research on Indian news media bias has largely focused on Indian English news outlets. For example, \cite{sen2022analysis} analyzed Indian English newspapers and their coverage of four large-scale news events to study information coverage bias and user reactions. Similarly, \cite{chauhan2024media} and \cite{rawat2022media} used sentiment-based analysis to understand bias in Indian English news sources.

\par Existing datasets for political bias detection in Indian languages are summarized in Table~\ref{tab:related_work}. One of the earliest works on political media bias detection in Indian languages is \cite{gangula2019detecting}. This study created a dataset of 1,329 news articles collected from multiple Telugu newspapers and annotated them for bias toward specific political parties. Each article was annotated by four native Telugu speakers with good language proficiency and political awareness. Articles were labeled as biased toward one of five placeholder parties (A, B, C, D, E) or as None if unbiased. Annotators were instructed to read articles and headlines as they would in a newspaper and then judge whether the content was biased for or against a party. Building on this work, \cite{agrawal2022towards} introduced a dataset of 1,388 Hindi news articles and headlines collected from various Hindi news outlets. The task involved classifying articles into three categories — for, neutral, and against - with respect to the ruling party, represented by a placeholder X. The study evaluated multiple models and found that XLM-RoBERTa \cite{conneau2020unsupervisedcrosslingualrepresentationlearning} performed best, achieving an accuracy of 83.00 \%, a macro-F1 score of 76.4\%, and a Matthews Correlation Coefficient of 72.1\%. However, a key limitation of these work are and the earlier Telugu study is that the datasets focus exclusively on the governing party based articles. A more robust and neutral approach would involve collecting articles without such filters and covering perspectives related to all political parties. Additionally, none of these works explore the plethora of different types of biases possible but rather focus on understanding the difference of title wit respect to news article contents (similar to only a particular type of bias, i.e., \textit{click bait})

\begin{table*}[!t]
\centering
\caption{Summary of Created Datasets on Political Bias Detection in Indian Languages}
\label{tab:related_work}
\renewcommand{\arraystretch}{1.2}
\resizebox{\textwidth}{!}{%
\begin{tabular}{p{3.8cm} p{2.1cm} p{3.5cm} p{3.8cm} p{4.0cm}}
\hline
\textbf{Paper} & \textbf{Size / Lang.} & \textbf{Description} & \textbf{Annotation} & \textbf{Experiments} \\
\hline

Detecting Political Bias in News Articles Using Headline Attention \cite{gangula2019detecting} 
& 1329 / Telugu
& Articles from multiple Telugu newspapers, labeled for bias toward one of five parties (BJP, Congress, TRS, TDP, YCP) or \textit{None}.
& 4 native Telugu speakers; disagreements resolved by discussion. Cohen's $\kappa = 0.90$.
& 5-fold CV; compared ML (Naive Bayes, SVM) and neural models (CNN, LSTM, GRU). Headline Attention Network achieved $\sim$89.5\% accuracy. \\
\hline

Enhancing Bias Detection Using Pragmatic Presupposition \cite{kameswari2020enhancing} 
& 1050 / Telugu
& 850 biased and 200 unbiased Telugu articles; bias treated as binary (present/absent).
& Sentence-level presupposition scoring (0/1); guidelines cover coreference, deixis, verbal suffixes, and rhetorical questions.
& 10-fold CV across Naive Bayes, LR, SVM, RF, MLP. MLP with presupposition features: accuracy $\sim$96.5\%, F1 $\sim$95.7. \\
\hline

Quantifying Magnitude of Political Bias (PoBiCo-21) \cite{kameswari2021towards} 
& 500 / Telugu
& Articles from \textit{Eenadu}, \textit{Andhrajyothy}, \textit{Sakshi}; 180 positive, 220 negative, 100 unbiased.
& 10 fine-grained bias techniques (e.g., amplification, whataboutism) labeled at direct, indirect, and headline levels. $\kappa = 0.715$.
& Bias magnitude quantified via sentiment-based technique ranking; IndicBERT for scoring. Correctly identified the more biased article in $\sim$72\% of pairs. \\
\hline

Towards Detecting Political Bias in Hindi News Articles \cite{agrawal2022towards} 
& 1388 / Hindi
& Political news from four outlets: \textit{The Wire} \& \textit{The Quint} (liberal) and \textit{OPIndia} \& \textit{The Frustrated Indian} (pro-BJP).
& 2 native speakers with opposing leanings; labels: \textit{pro-BJP}, \textit{anti-BJP}, \textit{neutral}. Resolved by third annotator; $\kappa = 0.65$.
& 10:1 train/validation split; metrics: accuracy, F1-macro, MCC. XLM-RoBERTa outperformed traditional ML. \\
\hline

Nishkarsh: AI-Based Multimodal Political Bias Detection \cite{bhojwani2025nishkarsh} 
& 13{,}000 / Hindi
& Large-scale dataset with headlines and full articles capturing pro/anti/neutral government stances across diverse outlets.
& Expert review combining manual checks with automated scoring (TF-IDF, embeddings).
& RoBERTa for leaning (87\% acc.), DistilBERT for sentiment (92.6\% acc.); outputs summarized via BART. \\
\hline

Read Between the Lines: Benchmark for Bangla News (BanglaBias) \cite{lia2025read} 
& 200 / Bangla
& $\sim$46 political events (2013–2025) from 54 outlets; 95 government-critique, 72 neutral, 33 government-leaning articles.
& 2 native speakers + adjudicator; structured decision flowchart; 3 stances. IAA: 73.5\% ($\kappa = 0.574$).
& Benchmarked across 28 LLMs (zero/few-shot). Larger models: F1 up to 0.83 for critique; neutral stance hardest. \\
\hline

\end{tabular}%
}
\end{table*}


\par \cite{kuila2024news} provides a detailed understanding of bias in Indian news through the lens of different events. However, they focus on understanding bias only at sentence level for English news articles, whereas we also intend to understand the implicit contextual bias at the news article level and for Indic languages. Additionally, Maity et al. \cite{maity2024multilingual} have provided a bias dataset for Wikipedia content in multiple Indian languages. Although the dataset is extensive, it is not directly applicable for news articles. Therefore, in order to bridge these research gaps, we propose a Hindi news article dataset considering $4$ different bias labels. Additionally, our annotators provide with \textit{rationale} which would be extremely helpful to build explainable automated media bias detection and mitigation approach.


\section{Dataset Overview} \label{s:data}

For our work, we have considered four leading Indian News Media agencies for Hindi Language, i.e., \textit{Aaj Tak}\footnote{\url{https://www.aajtak.in/}},  \textit{Live Hindustan}\footnote{\url{https://www.livehindustan.com/}}, \textit{Dainik Bhaskar}\footnote{\url{https://www.bhaskar.com/}} and \textit{Jagran}\footnote{\url{https://www.jagran.com/}}respectively. These four news publishers were selected because they have some of the highest audience reach and viewership among Hindi news media outlets. However, this claim cannot be supported by a single citation, as there is no official or centralized metric for ranking all news publishers across different media platforms. For \textit{Live Hindustan}, articles were randomly sampled using a Selenium-based scraping script applied to the website’s XML sitemap. For \textit{Aaj Tak}, \textit{Dainik Bhaskar} and \textit{Jagran}, articles were scraped from the most recent political news. Since the website does not provide an XML sitemap, random sampling of articles from previous months or years was not possible. Therefore, we implemented a JavaScript-based scraping script that extracts all article URLs from the political news category by identifying relevant \textit{<a>} tags. The script then filters out articles with word counts below a predefined threshold, removes duplicate entries, and stores the cleaned articles in a text file. These articles are subsequently processed and incorporated into the original CSV dataset. For pre-processing, duplicate records were removed to avoid redundancy, and annotation labels were checked for consistency.  

\par An article could have either of these biases and not limited to only one bias type, such as, \textit{selection, framing, omissiom} and \textit{clickbait} biases. We provide a brief description of each of these biases with examples in Table \ref{tab:tab3}. After dataset collection, we organize our dataset into following fields, \textit{news\_publisher} which comprises of the name, \textit{article\_date}, \textit{article\_links} which contains the link to the news article, and label information that indicates whether any form of bias is present in the article and separate columns for each of the $4$ types of bias along with \textit{Explanation} that includes the rationale provided by 3 annotators who labeled a particular article. We next discuss the annotation procedure. Further statistical information about the dataset can be found in Table~\ref{tab:dataset-statistics} and Table~\ref{tab:tab4}.

\begin{table*}[!t]
\centering
\caption{Examples of Explicit Language-Based Media Bias Categories}
\label{tab:tab3}
\small
\begin{tabular}{|p{3cm}|p{7cm}|p{7cm}|}
\hline
\textbf{Bias Labels} & \textbf{Examples (Hindi)} & \textbf{Translated (English)} \\
\hline\hline

\textbf{1. Framing Bias} &
\textbullet~\textbf{Example 1:} \texthindi{राजनीतिक दल मिलकर बिहार विधानसभा चुनाव में मतदाता सहभागिता बढ़ाने की रणनीति बना रहे हैं।}\par
\textbullet~\textbf{Example 2:} \texthindi{दिल्ली विधानसभा चुनाव में} A \texthindi{के केंद्रीय नेतृत्व ने पार्टी की अंदरूनी गुटबाजी पर लगाम लगाई है।}
&
\textbullet~\textbf{Example 1:} Political parties are coming together to strategize on increasing voter participation in the Bihar Assembly elections.\par
\textbullet~\textbf{Example 2:} In the Delhi Assembly elections, the central leadership of the A has curbed the internal factionalism within the party.
\\
\hline


\textbf{2. Clickbait Bias} &
\textbullet~\textbf{Example 1:} \texthindi{धनबाद–चंद्रपुरा पैसेंजर के बहाने} D \texthindi{का सियासी मास्टर स्ट्रोक}\par
\textbullet~\textbf{Example 2:} \texthindi{राजनीतिक दलों को आदर्श आचार संहिता का पाठ पढ़ाया गया।}
&
\textbullet~\textbf{Example 1:} D's political masterstroke under the pretext of Dhanbad–Chandrapura passenger.\par
\textbullet~\textbf{Example 2:} Political parties were taught the code of conduct.
\\
\hline

\end{tabular}
\end{table*}

\begin{table}[!t]
    \centering
     \caption{Dataset Statistics}
    \begin{tabular}{|l|c|c|}
    \hline
    \textbf{} & \textbf{Biased} & \textbf{Unbiased} \\
    \hline
    \textbf{Articles} & 240 & 208 \\
    \hline
    \textbf{Avg \#words in headline} & 15.40 & 14.81 \\
    \hline
    \textbf{Avg \#words in article} & 682.60 & 521.62 \\
    \hline
    \textbf{Avg \#sentences in article} & 19.69 & 16.88 \\
    \hline
    \end{tabular}
    \label{tab:dataset-statistics}
\end{table}

\begin{table*}[!t]
\centering
\caption{News Publisher Statistics}
\label{tab:tab4}
\begin{tabular}{|l|c|l|l|}
\hline
\textbf{News Publisher} & \textbf{Number of Articles} & \textbf{Biased/Unbiased Articles} & \textbf{Date Range} \\
\hline
\multirow{2}{*}{1. Live Hindustan} & \multirow{2}{*}{171} & Biased: 89 & \multirow{2}{*}{Jan 11 -- Nov 14, 2024} \\
\cline{3-3}
 & & Unbiased: 82 & \\
\hline
\multirow{2}{*}{2. Aaj Tak} & \multirow{2}{*}{110} & Biased: 68 & \multirow{2}{*}{Nov 28 -- Dec 27, 2025} \\
\cline{3-3}
 & & Unbiased: 42 & \\
\hline
\multirow{2}{*}{3. Dainik Bhaskar} & \multirow{2}{*}{68} & Biased: 27 & \multirow{2}{*}{Dec 21 -- Dec 29, 2025} \\
\cline{3-3}
 & & Unbiased: 41 & \\
\hline
\multirow{2}{*}{4. Jagran} & \multirow{2}{*}{99} & Biased: 56 & \multirow{2}{*}{Dec 12 -- Dec 27, 2025} \\
\cline{3-3}
 & & Unbiased: 43 & \\
\hline
\end{tabular}
\end{table*}
\subsection{Data Annotation} 
\label{subsec:data_anno}
A total of $14$ annotators were involved in the creation of this dataset, and each article has been annotated by $3$ experts. We initially started creating this dataset with $9$ bias categories: clickbait bias, selection bias, framing bias, omission bias, factual bias, linguistic bias, emotional bias, stance bias, and ideology bias. Upon analyzing the distribution of each bias across $200$ news articles, we limited our study to the top four—clickbait bias, selection bias, framing bias, and omission bias—as illustrated in Figure~\ref{fig:dummy_fig2}. These represented the most relevant biases according to frequency of occurence and therefore, seemed essential for future work involving the development of media bias mitigation models. We followed the following conditions for annotation : 

\begin{figure*}[!t]
    \centering
    \includegraphics[width=0.7\textwidth]{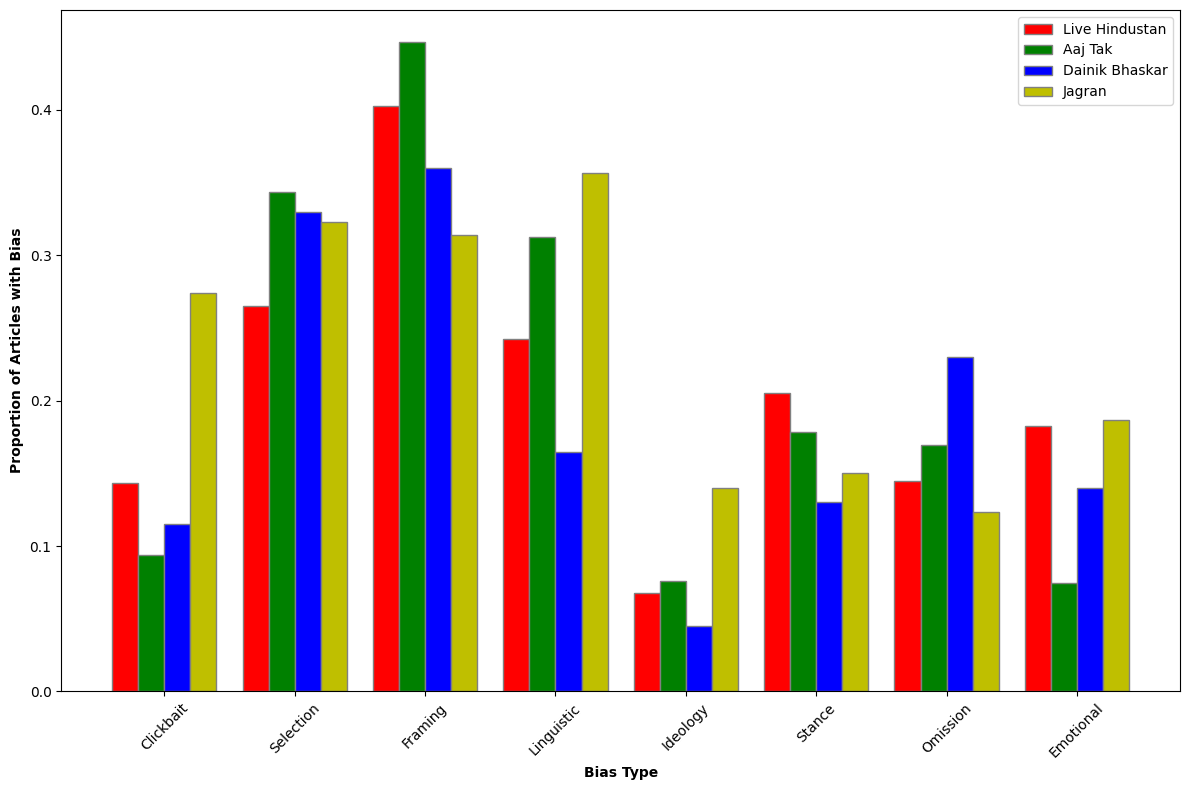}
    \caption{Distribution of fine-grained bias categories across news agencies.}
    \label{fig:dummy_fig2}
\end{figure*}

\begin{enumerate}
    \item All annotators were confirmed to be politically neutral, i.e., they had no political incentives or inclinations toward any particular political party, political figure, or political ideology. This was ensured by assessing the responses of candidate annotators to a series of questions asked during in-person interviews conducted by the first authors. Additionally, each annotator was given a sample of $10$ news articles to annotate, and a discussion meeting was held to resolve any ambiguity, given that bias annotations can often be subjective in nature. Candidate selection was strictly determined by the outcomes of the first two assessment phases described above.

    \item All annotators were provided with detailed training on the different types of biases using a comprehensive rulebook, as summarized in Table~\ref{tab:tab2}. It was ensured that each annotator complied to these two factors before annotation procedure. Additionally, we provide a detailed documentation of each type of biases considered in the dataset documentation. While there are some examples that are language specific and require understanding of the language specific information, such as Hindi, the remaining biases require understanding of the logical structure and reasoning of the article. We highlight a few examples of biases that are explicitly classified based on the language of the article in Table~\ref{tab:tab3}. The annotators comprised information technology students, working professionals, and professors from different Indian states who possess native or proficient multilingual capabilities in both Hindi and their respective state languages. 
\end{enumerate}

\begin{table*}[!t]
\centering
\caption{Annotation Guidelines for Bias Classification}
\label{tab:tab2}
\footnotesize
\renewcommand{\arraystretch}{1.15}
\setlength{\tabcolsep}{4pt}
\begin{tabularx}{\textwidth}{|L{2.4cm}|X|}
\hline
\textbf{Bias Label} & \textbf{Description} \\
\hline\hline

\textbf{1. Selection Bias} &
\textbf{Rule:} Focus on what is included or, more importantly, excluded from the story or the entire news cycle.

\textbullet~\textbf{Inclusion/Exclusion of Sources:} A story predominantly features sources supporting one side (e.g., only Party X spokespersons) while ignoring or minimizing opposing voices (e.g., opposition leaders).

\textbullet~\textbf{Topic Filtering:} One key political event, scandal, or achievement related to one party is given zero coverage or is buried deep, while similar news about another party is highly publicized.

\textbullet~\textbf{Data Cherry-picking:} Using only the specific statistics or facts (e.g., economic growth rates, poll numbers) that favor a particular narrative and omitting inconvenient data. \\
\hline

\textbf{2. Framing Bias} &
\textbf{Rule:} Focus on how the story is packaged, emphasizing certain aspects to influence interpretation.

\textbullet~\textbf{Choice of Metaphor/Analogy:} Using loaded terms or imagery to describe a party or event (e.g., describing a protest as `chaos/anarchy' (\texthindi{अराजकता}) vs.\ `democratic expression' (\texthindi{लोकतांत्रिक अभिव्यक्ति})).

\textbullet~\textbf{Order of Information:} Placing the preferred narrative's conclusion or preferred facts (e.g., ``The government succeeded...'') at the very start of the headline/story.

\textbullet~\textbf{Focus of Responsibility:} Attributing success/positive outcomes to one party's `vision' (\texthindi{दृष्टि}) or `leadership' (\texthindi{नेतृत्व}), but blaming negative outcomes on external factors or the `past' (\texthindi{अतीत}). \\
\hline

\textbf{3. Linguistic Bias} &
\textbf{Rule:} Focus on the specific words, adjectives, and nouns used to describe people, groups, or actions.

\textbullet~\textbf{Loaded Adjectives:} Describing an action using emotionally charged words (e.g., a leader's firm decision is called `bold' (\texthindi{साहसिक}) vs.\ `stubborn' (\texthindi{अड़ियल})).

\textbullet~\textbf{Labeling:} Using different labels for similar groups (e.g., calling one group's supporters `patriots' (\texthindi{देशभक्त}) and an opposing group's supporters `agitators' (\texthindi{उपद्रवी})).

\textbullet~\textbf{Euphemisms/Dysphemisms:} Softening negative actions (e.g., calling a major financial collapse a `minor adjustment' (\texthindi{मामूली फेरबदल})) or hardening neutral ones. \\
\hline

\textbf{4. Ideology Bias} &
\textbf{Rule:} Focus on how the news story aligns with or subtly promotes a specific political, economic, or social doctrine.

\textbullet~\textbf{Normalization of Views:} Presenting a partisan or extreme ideological position (e.g., hyper-nationalism (\texthindi{अति राष्ट्रवाद}) or far-left policy (\texthindi{वामपंथी नीति})) as the only acceptable or mainstream viewpoint.

\textbullet~\textbf{Policy Critique:} Consistent, unquestioning support for or dismissal of policies based solely on the party's ideological stance (e.g., always praising privatization (\texthindi{निजीकरण}) or always condemning it).

\textbullet~\textbf{Historical Interpretation:} Re-interpreting historical events or figures in a way that aligns with the favored ideological narrative (\texthindi{वैचारिक कथा}) of a specific political base. \\
\hline

\textbf{5. Stance Bias} &
\textbf{Rule:} Focus on the reporter, anchor, or publication's express or implied approval or disapproval of a political actor or event.

\textbullet~\textbf{Approval/Disapproval in Tone:} The overall tone (\texthindi{स्वर}) of the narration is visibly fawning/respectful (\texthindi{प्रशंसात्मक/आदरपूर्ण}) when discussing one side and skeptical/derisive (\texthindi{संदेहपूर्ण/उपहासपूर्ण}) when discussing the other.

\textbullet~\textbf{Interviewer Behavior:} An anchor interrupts one interviewee frequently with leading questions but allows the opposing interviewee to speak uninterrupted.

\textbullet~\textbf{Visual/Audio Cues (in TV News):} Using flattering graphics/lighting for one leader and unflattering or aggressive visuals for an opponent. \\
\hline

\textbf{6. Omission Bias} &
\textbf{Rule:} Focus on the intentional leaving out of crucial context or necessary facts that would alter the story's meaning.

\textbullet~\textbf{Missing Context:} Reporting a politician's controversial statement without including the preceding question or the full sequence of events that provoked the comment.

\textbullet~\textbf{Withholding Key Facts:} Reporting on a crime rate increase in a specific state but omitting the fact that the reporting methodology for crimes changed, skewing the numbers.

\textbullet~\textbf{Incomplete Follow-up:} Failing to follow up on allegations or promises made by one side that turned out to be false or unfulfilled. \\
\hline

\textbf{7. Fact Bias} &
\textbf{Rule:} Focus on the use of verifiable information, particularly where a genuine fact is mixed with opinion or presented as fact.

\textbullet~\textbf{Opinion as Fact:} Presenting a claim that requires proof (e.g., ``The economy will grow by 10\%'') as a certain, established fact (\texthindi{स्थापित तथ्य}) without attribution or supporting evidence.

\textbullet~\textbf{Misrepresentation:} Distorting or twisting an actual quote, number, or event to support the desired narrative.

\textbullet~\textbf{Over-simplification:} Reducing complex policy issues to overly simplistic, easily digestible slogans that ignore critical details or caveats. \\
\hline

\textbf{8. Emotional Bias} &
\textbf{Rule:} Focus on language and presentation designed to elicit a strong emotional response (anger, fear, solidarity) rather than rational thought.

\textbullet~\textbf{Appeal to Fear/Anger:} Using language/imagery to suggest an opposing group or policy is an existential threat (\texthindi{अस्तित्व का ख़तरा}) to the nation, religion, or society.

\textbullet~\textbf{Appeal to Pity/Sympathy:} Over-dramatizing the plight of one group (often victims of the `other side') to elicit uncritical sympathy (\texthindi{सहानुभूति}) and support.

\textbullet~\textbf{Use of Anecdotes:} Relying solely on a highly emotive, singular personal story (\texthindi{भावनात्मक व्यक्तिगत कहानी}) to represent a complex trend or policy outcome, overriding statistical evidence. \\
\hline

\textbf{9. Clickbait Bias} &
\textbf{Rule:} Focus on the headline. Assign this label if the headline is designed to attract readers through sensational, controversial, or misleading statements, or by selectively highlighting a single line or detail---often taken out of context or unrelated to the main focus of the article---without accurately representing the overall content. \\
\hline

\end{tabularx}
\end{table*}

\par We ensured that each article was annotated by three annotators, with disagreements resolved through majority voting. However, individual annotators’ explanations for their decisions are retained in the dataset to enable computational models to incorporate human reasoning and to identify and mitigate bias in the articles. All articles were evenly distributed among the annotators and annotated across all four news agencies. The following protocols were enforced: (1) article assignments were randomly sampled, ensuring that no annotator annotated the same article more than once; and (2) each article was independently annotated without any prior knowledge of the annotations assigned by another annotator.

\par
\begin{figure*}[!t]
    \centering
    \includegraphics[scale=0.6]{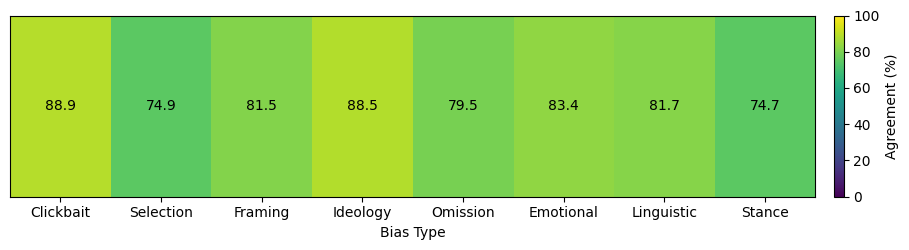}
    \caption{Inter-annotator agreement (\%) across all news agencies and bias types.}
    \label{fig:dummy_fig3}
\end{figure*}

 \textbf{Inter-Annotator Agreement} :  In order to ensure that the dataset is balanced with different annotator perspectives, we assess annotation reliability using Cohen's Kappa \cite{warrens2015five}.  For the binary biased versus non-biased  classification, we obtained a Kappa value of 0.81 (a news article is classified as biased if even one of the four bias categories is marked as present), indicating strong inter-annotator agreement. 
For the multi-label task, the Cohen’s Kappa values are 0.89, 0.75, 0.81, and 0.79 for clickbait, 
selection, framing, and omission biases, respectively, resulting in an average Kappa of 0.80, which 
reflects strong agreement. A per-agency, per-category breakdown of the agreement scores is presented in Figure~\ref{fig:dummy_fig3}.

\subsection{Summary of Insights}

We highlight our observations during dataset creation considering the different types of biases possible next :
\begin{itemize}

    \item It should be noted that, despite the availability of a comprehensive rulebook for the classification of bias categories into binary labels, a degree of inherent subjectivity remains unavoidable. This is particularly evident in bias categories such as emotional bias, where different annotators may interpret the tone, implications, or neutrality of an article differently. Similarly, omission bias is inherently difficult to identify, as assessing whether important information has been deliberately excluded often requires broader contextual knowledge of the event beyond what is presented in the article.

    \item The biases initially studied before narrowing down to four are not mutually exclusive; multiple biases may be present in a single article, and some may act as subsets of others. Furthermore, subtle distinctions exist between labels that might otherwise appear similar. For example, linguistic bias depends exclusively on the language used in the article and may imply emotional bias (see Table~\ref{tab:linguistic-bias}). However, emotional bias can also arise from the overall tone of reporting even when explicit linguistic markers, such as idioms or offensive words, are absent (see Table~\ref{tab:emotional-bias}). A similar subtle distinction also occurs between selection bias and omission bias (see Table~\ref{tab:selection-bias} and Table~\ref{tab:omission-bias}).

    \item It is a common observation that many articles quote statements made by political figures. We report only the bias that arises from the author’s reporting and not from the political figure’s speech, as our focus is on media bias detection and understanding. However, in cases where an article captures only a single figure’s speech or repeatedly emphasizes a single controversial statement, bias may be introduced by the author of the news article, thereby making the article biased.
\end{itemize}



\begin{table*}[!t]
\centering
\small
\renewcommand{\arraystretch}{1.4}
\caption{Example of \textbf{Linguistic Bias} without Emotional Bias. Loaded terms shape political framing, but the text does not attempt to manipulate the reader's emotional response against any party.}
\label{tab:linguistic-bias}
\begin{tabular}{p{2.2cm} p{11.8cm}}
\toprule
\textbf{Field} & \textbf{Content} \\
\midrule

\textbf{Title} &
\texthindi{खरमास बाद बिहार में आएगा सियासी भूचाल?} \textbf{A} \texthindi{का दावा,} \textbf{B C} \texthindi{को देंगे झटका} \\[4pt]

\textbf{Article (excerpt)} &
\texthindi{बिहार के पूर्व मुख्यमंत्री और दल M के सुप्रीमो} \textbf{A} \texthindi{ने} \textbf{X} \texthindi{को लेकर बड़ा दावा किया है। कहा है कि खरमास के बाद} \textbf{B} \texthindi{बड़ा फैसला लेने वाले हैं। वह कभी भी} \underline{\texthindi{पलटी मार}} \texthindi{कर} \textbf{C} \texthindi{और} \textbf{X} \texthindi{के दलों को} \underline{\texthindi{झटका}} \texthindi{दे सकते हैं।} \ldots \textbf{A} \texthindi{ने कहा कि} \textbf{B} \texthindi{पर} \textbf{C} \texthindi{का} \underline{\texthindi{दबाव}} \texthindi{है।} \ldots \texthindi{ऐसे में एक विकल्प यह भी है कि} 14 \texthindi{जनवरी के बाद या तो वह} \underline{\texthindi{पलटी मारेंगे}} \texthindi{या विधानसभा भंग कर देंगे।}
\\[4pt]

\textbf{English Summary} &
Former state CM \textbf{A} claimed that current CM \textbf{B} is poised to deliver a major blow to \textbf{C}'s political alliance \textbf{X} by potentially switching back to alliance \textbf{Y}. \textbf{A} stated that \textbf{B} is ready to rejoin the \textbf{Z}-led alliance due to discomfort over corruption allegations surrounding Deputy CM \textbf{D}, alongside growing pressure from party \textbf{P} to hand over state leadership. \textbf{B} has a historical pattern of shifting alliances and will likely either flip sides or dissolve the assembly rather than step down. \\[4pt]

\textbf{Linguistic Bias} &
\textbf{Present.} Phrases such as \underline{\texthindi{''पलटी मारेंगे''}} (\textit{will flip/turn around}), \underline{\texthindi{''दबाव''}} (\textit{pressure}), and \underline{\texthindi{''झटका''}} (\textit{jolt/blow}) are loaded political terms introduced by the reporter that cast one alliance in an unfair light, introducing framing bias through word choice. \\[4pt]

\textbf{Emotional Bias} &
\textbf{Absent.} The text does not attempt to evoke strong feelings or manipulate the reader's sentiments against any party. The narrative remains descriptive and factual without an affective charge. \\

\bottomrule
\end{tabular}
\end{table*}


\begin{table*}[!t]
\centering
\small
\renewcommand{\arraystretch}{1.4}
\caption{Example of \textbf{Emotional Bias} without Linguistic Bias. Unchallenged reporting of inflammatory accusations may agitate certain reader groups, yet the reporter's own word choices remain neutral.}
\label{tab:emotional-bias}
\begin{tabular}{p{2.2cm} p{11.8cm}}
\toprule
\textbf{Field} & \textbf{Content} \\
\midrule

\textbf{Title} &
\textbf{Z} \texthindi{की सरकार वोट एवं तुष्टिकरण की कर रही राजनीति:} \textbf{A} \\[4pt]

\textbf{Article (excerpt)} &
\texthindi{राष्ट्रीय अध्यक्ष} \textbf{A} \texthindi{ने कहा वर्तमान सत्ताधारी सरकार सामान्य जन को भयभीत करने के लिए लोगों को बिना किसी उचित कारण के जेलों में बंद कर रही है। वर्तमान सरकार विशेष रूप से मुसलमान, दलितों, पिछड़ों एवं आदिवासियों का उत्पीड़न कर रही है। किसानों के साथ सौतेला व्यवहार किया जा रहा है। यह दुर्भाग्य की बात है कि देश की हर सियासी पार्टियां मुसलमान के साथ सौतेला व्यवहार कर रही है।}
\\[4pt]

\textbf{English Summary} &
An ulema conference of party \textbf{X} was organized in Pachperwa, where national president \textbf{A} strongly criticized the ruling \textbf{Z} government, labeling it anti-farmer and accusing it of practicing vote-bank and appeasement politics. \textbf{A} alleged that the administration is unjustly imprisoning citizens to create fear and is exploiting marginalized groups including Muslims, Dalits, backward classes, and tribals. He concluded that if party \textbf{X} comes to power, its leaders will fight impartially for the justice of all communities. \\[4pt]

\textbf{Emotional Bias} &
\textbf{Present.} The article carries a mild emotional bias by solely and uncritically reporting the inflammatory accusations of party \textbf{X}---allegations of unjust imprisonment, persecution of minorities, and exploitation of marginalized communities. Presenting these charged claims without any counter-perspective may provoke strong negative emotions toward the \textbf{Z} government among certain reader groups. \\[4pt]

\textbf{Linguistic Bias} &
\textbf{Absent.} The reporter does not employ loaded or partisan language of their own. All critical statements are straightforwardly attributed to \textbf{A}. No reporter-level framing terms appear; the language remains neutral at the lexical level. \\

\bottomrule
\end{tabular}
\end{table*}


\begin{table*}[!t]
\centering
\small
\renewcommand{\arraystretch}{1.4}
\caption{Example of \textbf{Selection Bias} without Omission Bias. The reporter selectively covers a single-party event, but within that chosen story, no critical context or counter-argument is strategically withheld.}
\label{tab:selection-bias}
\begin{tabular}{p{2.2cm} p{11.8cm}}
\toprule
\textbf{Field} & \textbf{Content} \\
\midrule

\textbf{Title} &
\texthindi{अवध ने यूपी ही नहीं पूरे देश को दिया है साफ राजनीति का संदेश,} \textbf{A} \texthindi{का निशाना} \\[4pt]

\textbf{Article (excerpt)} &
\textbf{Q} \texthindi{महासचिव} \textbf{A} \texthindi{ने मंगलवार को रायबरेली में कहा कि कार्यकर्ताओं ने कठिन से कठिन परिस्थितियों में काम किया।} \ldots \texthindi{अवध से पूरे देश में संदेश गया कि हमें समर्पित, सच्ची और साफ राजनीति चाहिए।} \ldots \textbf{F} \texthindi{ने कहा कि देश में} 2014 \texthindi{में राजनीतिक बदलाव आया। यह बदलाव सिर्फ} 20-25 \texthindi{लोगों के लिए रहा। देश में पहले नफरत फैलाते हैं और अरबपतियों को फायदा पहुंचाते हैं।}
\\[4pt]

\textbf{English Summary} &
After the electoral victory in Rae Bareli, leaders \textbf{A} and \textbf{F} of alliance \textbf{I} thanked workers and voters. \textbf{A} credited collective efforts in Amethi and Rae Bareli and called for clean politics. MP \textbf{F} criticized leader \textbf{M} and the ruling administration for division, favoring billionaire \textbf{G}, and neglecting unemployment and inflation, and pledged to oppose the Agniveer scheme. \\[4pt]

\textbf{Selection Bias} &
\textbf{Present.} The reporter chose to cover exclusively a victory rally held by alliance \textbf{I}, presenting only their speeches, internal dynamics, and criticisms of the ruling party. The bias lies at the \emph{event-selection} level: by picking this one-sided event, the coverage structurally centers a single alliance's narrative. \\[4pt]

\textbf{Omission Bias} &
\textbf{Absent.} Within the scope of the chosen event, the reporter does not suppress any critical facts or mandatory context. The story faithfully covers what transpired at the rally; no distortion through selective exclusion of relevant details occurs \emph{within} the reported event itself. \\

\bottomrule
\end{tabular}
\end{table*}


\begin{table*}[!t]
\centering
\small
\renewcommand{\arraystretch}{1.4}
\caption{Example of \textbf{Omission Bias} without Selection Bias. The reporter covers a legitimate, newsworthy political event but strategically withholds the incumbent's response and record, skewing the narrative within the story itself.}
\label{tab:omission-bias}
\begin{tabular}{p{2.2cm} p{11.8cm}}
\toprule
\textbf{Field} & \textbf{Content} \\
\midrule

\textbf{Title} &
\texthindi{जमशेदपुर पूर्वी में पूर्व} \textbf{A} \texthindi{अध्यक्ष} \textbf{X} \texthindi{ने शुरू किया जनसंपर्क अभियान} \\[4pt]

\textbf{Article (excerpt)} &
\ldots \texthindi{पूर्व मुख्यमंत्री} \textbf{Z} \texthindi{के भांजे व पूर्व महानगर} \textbf{A} \texthindi{अध्यक्ष} \textbf{X} \texthindi{ने जन संपर्क अभियान शुरू कर दिया।} \ldots \texthindi{बस्तीवासियों ने विधायक} \textbf{Y} \texthindi{की कार्यशैली की आलोचना की।} \textbf{X} \texthindi{को जनसमूह ने बताया कि क्षेत्र की समस्याएं गंभीर हैं} \ldots \texthindi{गंदगी का अंबार, सार्वजनिक शौचालय का हाल बेहाल, बजबजाती हुई नालियां चारों ओर नजर आ रही हैं, इसके अलावे बस्ती की मूलभूत सुविधाओं को लेकर भी विधायक संवेदनशील नहीं हैं।}
\\[4pt]

\textbf{English Summary} &
Ahead of the Jharkhand assembly elections, former metropolitan \textbf{A} chief \textbf{X}, nephew of former CM \textbf{Z}, began a public outreach campaign in Jamshedpur East. Residents at a Jojobera meeting criticized sitting MLA \textbf{Y}'s performance, citing garbage, poorly maintained toilets, overflowing drains, and inadequate basic facilities. No response, counter-claims, or achievements from \textbf{Y} are presented. \\[4pt]

\textbf{Omission Bias} &
\textbf{Present.} The reporter covers the broader political story but, within it, strategically omits the incumbent MLA \textbf{Y}'s response, perspective, and any positive developmental record. This selective exclusion of critical context within an already covered topic skews the reader's perception in favor of the challenger---a hallmark of omission bias. \\[4pt]

\textbf{Selection Bias} &
\textbf{Absent.} The reporter does not engage in gatekeeping-level bias by ignoring the story or choosing only one side's event. Political competition ahead of the Jharkhand elections is a legitimate and newsworthy topic; the bias lies not in \emph{which story} was selected, but in \emph{what was left out} within that story. \\

\bottomrule
\end{tabular}
\end{table*}

\section{Analysis and Observations} \label{s:res}
\par
\begin{figure*}[!t]
    \centering
    \hspace*{-1cm}
    \includegraphics[scale=0.45]{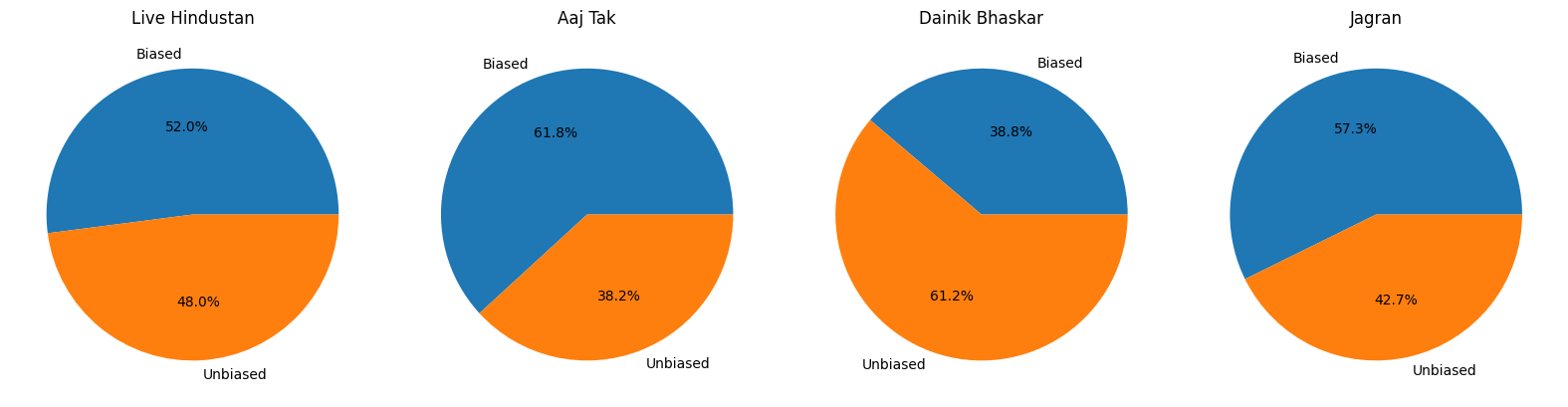}
    \caption{Distribution of biased and unbiased news articles across agencies.}
    \label{fig:dummy_fig1}
\end{figure*}

 In this Section, we discuss our observations based on the annotated dataset with respect to the leading news media agencies that publish news in Hindi. Our initial observations include a visualization of the proportion of biased news articles for each news media agency, which aids in understanding the exposure of the Hindi news reading population to biased reporting. We observe that a substantial fraction of Hindi political news articles exhibit bias. Specifically, $52\%$, $62\%$, $40\%$ and $57\%$ of the articles in our dataset from Live Hindustan, Aaj Tak, Dainik Bhaskar and Jagran are biased, as shown in Table~\ref{tab:tab4}. Figure~\ref{fig:dummy_fig1} provides a visualization of the bias prevalence across the agencies, where the biased share clearly dominates or is comparable to the unbiased share in both cases. While we acknowledge that these proportions are based on a sampled subset of articles, they nonetheless indicate that a large number of Hindi news articles, irrespective of the media agency, carry some form of bias. This further underscores the need for a large-scale media bias dataset for the Hindi language to enable automated bias detection.

\subsection{Case Study : Linguistic and Structural Feature Analysis}
\par Beyond the label-level statistics, we conducted an exploratory analysis to examine whether biased and unbiased articles differ along measurable surface-level, lexical, and syntactic dimensions. Understanding these differences is important for two reasons: it helps characterize \textit{how} bias manifests linguistically in Hindi political reporting, and it establishes whether shallow lexical cues alone are sufficient for automated detection or whether deeper contextual reasoning is required. For this analysis, we constructed a feature-extraction pipeline that operates entirely on Devanagari-aware tokenization using the Indic NLP Library \cite{kunchukuttan2020indicnlp}, thereby avoiding the loss of nuance that arises from translation-based approaches.

\par The pipeline computes a broad set of article-level features spanning several linguistic dimensions. Lexical and part-of-speech features were derived using IndoWordNet \cite{bhattacharyya-2010-indowordnet}, from which we drew coverage lexicons of $19{,}417$ adjectives, $82{,}747$ nouns, and $2{,}212$ adverbs, supplemented with curated domain-specific political vocabulary and a hand-verified stopword list of $846$ entries built on top of the LiHiSTO \cite{rajwal2024lihisto} resource. We further computed sentiment polarity using curated positive and negative lexicons, superlative and intensifier density, article- and sentence-length statistics, lexical diversity measures such as the type-token ratio and Yule's characteristic constant $K$, and Hindi-specific readability estimates based on the RH1 and RH2 formulations. Finally, we obtained syntactic-complexity features - mean dependency distance, parse-tree depth, subject/object/clause ratios, and a composite complexity score by parsing every sentence with the Stanza Universal Dependencies pipeline for Hindi. In total, this yields $130$ features per article, which we release alongside the dataset to support future feature-driven studies.

\par Among these measures, sentence-level lexical diversity, quantified by Yule's $K$, proved most informative for distinguishing certain bias types, and we therefore report it in detail. Since Yule's $K$ is not normally distributed, we assess group differences using the non-parametric Mann-Whitney $U$ test, which avoids the skewness of standard parametric tests by evaluating rank-based medians instead of means. Table~\ref{tab:yulesk-bias} summarizes the median sentence-level Yule's $K$ for articles exhibiting each fine-grained bias against those that do not. We observe that, for the overall biased versus unbiased partition, there is no statistically significant difference in either the mean ($p=0.9951$) or the dispersion ($p=0.5565$) of sentence-level lexical richness. A similar pattern holds for framing ($p=0.5836$), selection ($p=0.9859$), and omission ($p=0.5239$) biases, none of which are separable from their respective negative classes on the basis of lexical diversity alone.

\begin{table*}[!t]
    \centering
    \caption{Sentence-level Yule's $K$ (median) for articles with versus without each bias type, and the corresponding Mann-Whitney $U$ $p$-value. $^{*}$ denotes significance at $p<0.05$.}
    \begin{tabular}{|l|c|c|c|c|}
    \hline
    \textbf{Bias Type} & \textbf{$n_{1}$} & \textbf{Median (Bias)} & \textbf{Median (No Bias)} & \textbf{$p$-value} \\
    \hline
    \textbf{Clickbait} & 45 & 69.61 & 62.89 & $0.0294^{*}$ \\
    \hline
    \textbf{Framing} & 99 & 63.52 & 65.59 & $0.5836$ \\
    \hline
    \textbf{Selection} & 67 & 64.24 & 64.77 & $0.9859$ \\
    \hline
    \textbf{Omission} & 32 & 58.70 & 65.72 & $0.5239$ \\
    \hline
    \end{tabular}
    \label{tab:yulesk-bias}
\end{table*}

\par The notable exception is clickbait bias, for which articles flagged as clickbait exhibit a significantly higher median sentence-level Yule's $K$ than non-clickbait articles ($69.61$ versus $62.89$, $p=0.0294$). This is intuitive: clickbait framing tends to rely on denser and more varied vocabulary - rhetorical hooks, sensational descriptors, and out-of-context teasers which inflates lexical richness relative to routine reporting. The corresponding dispersion of Yule's $K$ trends in the same direction, though it does not reach significance ($p=0.1844$).


\par Taken together, these findings carry an important implication. With the sole exception of clickbait - whose signal is largely superficial and headline-driven - the fine-grained bias categories that dominate our dataset, namely selection, framing, and omission, are \textit{not} distinguishable from unbiased articles using surface lexical-diversity statistics. In other words, an article that omits crucial context or selectively frames an event is written in language that is, statistically, indistinguishable from neutral reporting. This corroborates our central premise that media bias in Hindi political news is predominantly a contextual and inferential phenomenon rather than a lexical one, and it is therefore unlikely to be captured by shallow feature-based classifiers. It also explains, at least in part, why the language models evaluated over-predict lexically driven labels (illustrated in Table~\ref{tab:llm-human-example-part1} and Table~\ref{tab:llm-human-example-part2}) while missing biases that require reasoning over what is included, foregrounded, or withheld. These observations reinforce the need for reasoning-oriented and culturally grounded models, and motivate the release of both the bias labels and the accompanying annotator rationales.


\begin{table*}[!t]
\centering
\footnotesize
\renewcommand{\arraystretch}{1.15}
\setlength{\tabcolsep}{5pt}

\caption{An example from the dataset illustrating LLM vs.\ human annotation differences across nine bias labels. Part I: News article, prompt, and LLM response.}
\label{tab:llm-human-example-part1}

\begin{tabular}{p{2.3cm} p{12.1cm}}
\toprule
\textbf{Field} & \textbf{Content} \\
\midrule

\textbf{News Article} &
\texthindi{वार पलटवार: दल A के नेता व्यक्ति A ने व्यक्ति B पर साधा निशाना, दल B नेता को बताया भारत के लोकतंत्र और संविधान पर सबसे बड़ा हमला}

\vspace{3pt}

\texthindi{डिजिटल डेस्क, नई दिल्ली। दल B सांसद व्यक्ति C द्वारा व्यक्ति D को प्रधानमंत्री बनाए जाने के बयान पर दल A नेता व्यक्ति A ने प्रतिक्रिया देते हुए तंज कसा है। उन्होंने कहा कि व्यक्ति C पहले व्यक्ति B से मिल लें और यही बयान एक बार व्यक्ति B की जुबान से दिलवा दें।} \ldots

\vspace{3pt}

\texthindi{व्यक्ति A ने कहा कि प्रधानमंत्री कोई सरनेम से नहीं बनता। उन्होंने व्यक्ति B को `लीडर ऑफ पर्यटन' बताते हुए कहा कि वह सदन की कार्यवाही छोड़कर दूसरे देश चले जाते हैं। भारत के लोकतंत्र और संविधान पर सबसे बड़ा हमला तो व्यक्ति B खुद करते हैं, जब वह लोकतंत्र को लेकर विदेश की धरती पर सवाल उठाते हैं।}
\\[6pt]

\midrule

\textbf{Prompt} &
[News Article] + Read this article carefully and classify the article into binary labels for each of the following media bias labels: 1.\ clickbait bias (for the title), 2.\ selection bias, 3.\ framing bias, 4.\ linguistic bias, 5.\ ideology bias, 6.\ stance bias, 7.\ omission bias, 8.\ fact bias, 9.\ emotional bias; also provide a brief explanation for why you believe a type of bias is present in the article, with reference/context from the article (if applicable).

\textit{[Note: The rulebook for classification of each bias was not prompted, to check the language model's intrinsic reasoning on the news article.]}
\\[6pt]

\midrule

\textbf{LLM Response} &
\textbf{Labels:} 0, 1, 1, 1, 1, 1, 1, 0, 1
\\[4pt]

& \textit{Selection Bias:} The article covers exclusively Party A leader Person A's press statements targeting Party B. There is no corresponding coverage of Party B's response, Person C's original statement in full context, or any rebuttal from the Party B side.
\\[4pt]

& \textit{Framing Bias:} The article is framed entirely as a Party A counter-offensive (\texthindi{वार पलटवार}), with Person A positioned as the active critic and Party B figures as passive targets, treating Party A's accusations as newsworthy claims without framing them as contested political rhetoric.
\\[4pt]

& \textit{Linguistic Bias:} Multiple loaded terms originating from Person A's speech are reproduced uncritically: \texthindi{तंज कसा} (sarcastically jabbed), \texthindi{सिर पर कुल्हाड़ी मार ली} (axed his own head), and \texthindi{लीडर ऑफ पर्यटन} (leader of tourism), normalizing a mocking register.
\\[4pt]

& \textit{Ideology Bias:} The article implicitly endorses a right-leaning position by uncritically amplifying Party A's anti-dynasty argument without ideological labeling or counterpoint.
\\[4pt]

& \textit{Stance Bias:} Person A's claims that Person B attacks democracy, lives off taxpayer money, and is a mere tourist are reported as substantive statements rather than partisan allegations, with no skepticism signaled.
\\[4pt]

& \textit{Omission Bias:} Person C's original statement is never quoted directly; Party B's response is absent; and no context is given for Person B's foreign visits.
\\[4pt]

& \textit{Emotional Bias:} The article reproduces emotionally charged language without journalistic distancing, likely provoking hostility toward Party B as a natural reading.
\\

\bottomrule
\end{tabular}
\end{table*}

\begin{table*}[!t]
\centering
\footnotesize
\renewcommand{\arraystretch}{1.15}
\setlength{\tabcolsep}{5pt}

\caption{An example from the dataset illustrating LLM vs.\ human annotation differences across nine bias labels. Part II: Gold labels and human justification.}
\label{tab:llm-human-example-part2}

\begin{tabular}{p{2.3cm} p{12.1cm}}
\toprule
\textbf{Field} & \textbf{Content} \\
\midrule

\textbf{Gold Labels} &
\textbf{Labels:} 1, 0, 1, 0, 0, 0, 0, 0, 0
\\[5pt]

& \textit{Clickbait Bias:} \texthindi{दल A के नेता व्यक्ति A ने व्यक्ति B पर साधा निशाना, दल B नेता को बताया भारत के लोकतंत्र और संविधान पर सबसे बड़ा हमला}
\\[4pt]

& \textit{Framing Bias:} Picks out the statements by Person A which are provocative, showing Party B leadership has been hierarchical but Party A leadership does not work that way, with no other arguments shown.
\\[6pt]

\midrule

\textbf{Justification} &
The article headline clearly tries to attract readers by showing how the Party A leader targets the Party B leader. However, the LLM confidently predicts ``there is no inflation and no misleading hook,'' raising no doubt about the strong and direct tone of the headline.

\vspace{4pt}

While selection bias could be debatable, it could alternatively be tagged as framing bias; where to mark it is considered subjective for this article. The reason manual annotators marked it as absent could be the lack of a response from Party B to this statement.

\vspace{4pt}

Furthermore, the LLM blunders by marking linguistic bias present; all the phrases it reports were spoken by the Party A leader, not the reporter. For linguistic bias to be present, these words or remarks should have been written by the article's editor.

\vspace{4pt}

Similarly, no specific ideology (leftist, rightist, or centrist) is promoted by the article, so ideological bias should be absent; however, the LLM identifies it anyway, providing a contradictory explanation.

\vspace{4pt}

While the reporter does not take a direct stand, one could mildly argue that the article favors Party A due to the lack of an opposition reply. Finally, omission bias is rightly marked as absent, as including Party B's reply would not significantly change the reader's perception of party support, given that the subject is relatively trivial.
\\

\bottomrule
\end{tabular}
\end{table*}

\section{FAIR Principles} \label{s:fair}
The dataset has been developed in accordance with the FAIR principles \cite{wilkinson2016fair} to ensure that it is legally and ethically fair and unbiased. We discuss each of these principles next.

\begin{itemize}
    \item \textbf{Findable:} The dataset is publicly available at \textbf{\url{https://github.com/KushalTrivedi19032005/Indic-Media-Bias-Detection-Dataset}}. Detailed metadata describing the dataset context, structure, and usage instructions is also provided. 
    
    \item \textbf{Accessible:} The dataset can be accessed through the above link and is publicly available to users.
    
    \item \textbf{Interoperable:} The dataset was originally created and annotated in CSV format, which is widely supported by data processing and analysis tools.
    
    \item \textbf{Reusable:} The dataset is hosted on a public platform that permits reuse for research purposes, subject to agreement with the stated terms and conditions.
\end{itemize}


\section{Conclusions} \label{s:con}
\par Media Bias can severely impact public perception of the event, form echo chambers, cause segregation and polarization in society. Therefore, identification of different forms of media bias and further mitigation and handling is severely required. Developing automated approaches for low-resource domains can be challenging due to the non-availability of task and domain-specific data. Additionally, understanding the implicit factors of Indian political context, specifically with the high variance across different languages is almost impossible. Therefore, we propose to create Hindi language specific large scale media bias benchmark datasets, that covers multiple facets of bias. In this paper, we presented a carefully curated annotated dataset for media bias detection in Hindi political news articles, addressing a significant gap in resources for low-resource domain for Indian context. The dataset consists of $448$ unique articles from four prominent Indian news agencies and is annotated by expert annotators across four fine-grained bias categories. By incorporating both bias labels and human-written explanations, the dataset supports not only bias detection but also explainable reasoning, which is crucial for advancing language technologies. Through an analysis of inter-annotator agreement, with Cohen's Kappa scores of $0.80$ across the four bias categories and $0.81$ for biased/unbiased annotations, we demonstrate that bias annotation is reliable when supported by clear guidelines. Our future directions include increase our dataset to ensure representation of different forms of demographic bias, such as, religion, region, language, gender, etc. This is specifically relevant to Indian context considering the high variance and diversity with respect to every demographic factor. We, additionally, believe that this fine-grained annotation would enable models to perform bias mitigation given a news article.

\section{Acknowledgment:} We would like to thank Mr. Murtuza Shaikh, Ms. Subhoshri Pal, Mr. Yash Kumar Panse, Mr. Vyom Tiwari, Mr. Rohit Kumawat, Ms. Anisha Khairnar, Mr. Satwik Duppala, Mr. Sai Pranav, Mr. Chakali Surendra, Ms. Mehr Aliah for their contribution in the annotation process.

\bibliography{bibliography}

\begin{thebibliography}{52}
\providecommand{\natexlab}[1]{#1}

\bibitem[{Agrawal et~al.(2022)Agrawal, Gupta, Gautam, and
  Mamidi}]{agrawal2022towards}
Agrawal, S.; Gupta, K.; Gautam, D.; and Mamidi, R. 2022.
\newblock Towards detecting political bias in hindi news articles.
\newblock In \emph{Proceedings of the 60th annual meeting of the association
  for computational linguistics: student research workshop}, 239--244.

\bibitem[{Anik et~al.(2025)Anik, Rahman, Wasi, and
  Ahsan}]{anik-etal-2025-preserving}
Anik, M.~A.; Rahman, A.; Wasi, A.~T.; and Ahsan, M.~M. 2025.
\newblock Preserving Cultural Identity with Context-Aware Translation Through
  Multi-Agent {AI} Systems.
\newblock In Truong, S.; Putri, R.~A.; Nguyen, D.; Wang, A.; Ho, D.; Oh, A.;
  and Koyejo, S., eds., \emph{Proceedings of the 1st Workshop on Language
  Models for Underserved Communities (LM4UC 2025)}, 51--60. Albuquerque, New
  Mexico: Association for Computational Linguistics.
\newblock ISBN 979-8-89176-242-8.

\bibitem[{Arifah et~al.(2025)Arifah, Maureen, Rofik, Puspila, Erifiawan
  et~al.}]{arifah2025social}
Arifah, I. D.~C.; Maureen, I.~Y.; Rofik, A.; Puspila, N. K.~W.; Erifiawan, H.;
  et~al. 2025.
\newblock Social Media Platforms in Managing Polarization, Echo Chambers, and
  Misinformation Risk in Interreligious Dialogue among Young Generation.
\newblock \emph{Journal of Social Innovation and Knowledge}, 1(aop): 1--33.

\bibitem[{Bhatia and Deepak(2018)}]{bhatia2018topic}
Bhatia, S.; and Deepak, P. 2018.
\newblock Topic-specific sentiment analysis can help identify political
  ideology.
\newblock In \emph{Proceedings of the 9th workshop on computational approaches
  to subjectivity, sentiment and social media analysis}, 79--84.

\bibitem[{Bhattacharyya(2010)}]{bhattacharyya-2010-indowordnet}
Bhattacharyya, P. 2010.
\newblock {I}ndo{W}ord{N}et.
\newblock In Calzolari, N.; Choukri, K.; Maegaard, B.; Mariani, J.; Odijk, J.;
  Piperidis, S.; Rosner, M.; and Tapias, D., eds., \emph{Proceedings of the
  Seventh International Conference on Language Resources and Evaluation
  ({LREC}'10)}. Valletta, Malta: European Language Resources Association
  (ELRA).

\bibitem[{Bhojwani et~al.(2025)Bhojwani, Shruti, Teli, Sharma, Hegde
  et~al.}]{bhojwani2025nishkarsh}
Bhojwani, P.; Shruti, J.; Teli, S.; Sharma, S.; Hegde, S.; et~al. 2025.
\newblock Nishkarsh: An AI-Based Multimodal Political Bias Detection System for
  Indian News Media.
\newblock In \emph{2025 International Conference on Emerging Technologies in
  Electronics and Green Energy (ICETEG)}, 1--8. IEEE.

\bibitem[{Chakraborty et~al.(2019)Chakraborty, Bhavsar, Dandapat, and
  Chandra}]{chakraborty2019tweet}
Chakraborty, R.; Bhavsar, M.; Dandapat, S.~K.; and Chandra, J. 2019.
\newblock Tweet summarization of news articles: An objective ordering-based
  perspective.
\newblock \emph{IEEE Transactions on Computational Social Systems}, 6(4):
  761--777.

\bibitem[{Chauhan et~al.(2024)Chauhan, Verma, Pandey, Srivastava, Malik, Garg,
  and Singh}]{chauhan2024media}
Chauhan, S.; Verma, S.; Pandey, A.; Srivastava, A.; Malik, P.; Garg, N.; and
  Singh, A.~K. 2024.
\newblock Media Bias Monitor: Quantifying Biases of Indian Print Media.
\newblock In \emph{2024 15th International Conference on Computing
  Communication and Networking Technologies (ICCCNT)}, 1--7. IEEE.

\bibitem[{Cheng, Durmus, and Jurafsky(2023)}]{cheng2023marked}
Cheng, M.; Durmus, E.; and Jurafsky, D. 2023.
\newblock Marked Personas: Using Natural Language Prompts to Measure
  Stereotypes in Language Models.
\newblock In \emph{Proceedings of the 61st Annual Meeting of the Association
  for Computational Linguistics (Volume 1: Long Papers)}, 1504--1532.

\bibitem[{Conneau et~al.(2020)Conneau, Khandelwal, Goyal, Chaudhary, Wenzek,
  Guzmán, Grave, Ott, Zettlemoyer, and
  Stoyanov}]{conneau2020unsupervisedcrosslingualrepresentationlearning}
Conneau, A.; Khandelwal, K.; Goyal, N.; Chaudhary, V.; Wenzek, G.; Guzmán, F.;
  Grave, E.; Ott, M.; Zettlemoyer, L.; and Stoyanov, V. 2020.
\newblock Unsupervised Cross-lingual Representation Learning at Scale.
\newblock arXiv:1911.02116.

\bibitem[{Entman(1993)}]{entman1993framing}
Entman, R.~M. 1993.
\newblock Framing: Towards clarification of a fractured paradigm.
\newblock \emph{McQuail's reader in mass communication theory}, 390: 397.

\bibitem[{Fan et~al.(2025)Fan, Lei, Corman, and Maciejewski}]{fan2025skeptik}
Fan, A.; Lei, F.; Corman, S.~R.; and Maciejewski, R. 2025.
\newblock Skeptik: A Hybrid Framework for Combating Potential Misinformation in
  Journalism.
\newblock \emph{ACM Transactions on Interactive Intelligent Systems}.

\bibitem[{Fan et~al.(2019)Fan, White, Sharma, Su, Choubey, Huang, and
  Wang}]{fan2019plain}
Fan, L.; White, M.; Sharma, E.; Su, R.; Choubey, P.~K.; Huang, R.; and Wang, L.
  2019.
\newblock In Plain Sight: Media Bias Through the Lens of Factual Reporting.
\newblock In \emph{Proceedings of the 2019 Conference on Empirical Methods in
  Natural Language Processing and the 9th International Joint Conference on
  Natural Language Processing (EMNLP-IJCNLP)}, 6343--6349.

\bibitem[{Gangula, Duggenpudi, and Mamidi(2019)}]{gangula2019detecting}
Gangula, R. R.~R.; Duggenpudi, S.~R.; and Mamidi, R. 2019.
\newblock Detecting political bias in news articles using headline attention.
\newblock In \emph{Proceedings of the 2019 ACL workshop BlackboxNLP: analyzing
  and interpreting neural networks for NLP}, 77--84.

\bibitem[{Garg, Chakraborty, and Dandapat(2025)}]{garg2025portrait}
Garg, P.~K.; Chakraborty, R.; and Dandapat, S.~K. 2025.
\newblock PORTRAIT: a hybrid aPproach tO cReate extractive ground-TRuth summAry
  for dIsaster evenT.
\newblock \emph{ACM Transactions on the Web}, 19(1): 1--36.

\bibitem[{Hamborg, Donnay, and Gipp(2019)}]{hamborg2019automated}
Hamborg, F.; Donnay, K.; and Gipp, B. 2019.
\newblock Automated identification of media bias in news articles: an
  interdisciplinary literature review.
\newblock \emph{International Journal on Digital Libraries}, 20(4): 391--415.

\bibitem[{Horych et~al.(2025)Horych, Mandl, Ruas, Greiner-Petter, Gipp, Aizawa,
  and Spinde}]{horych2025promises}
Horych, T.; Mandl, C.; Ruas, T.; Greiner-Petter, A.; Gipp, B.; Aizawa, A.; and
  Spinde, T. 2025.
\newblock The promises and pitfalls of LLM annotations in dataset labeling: A
  case study on media bias detection.
\newblock In \emph{Findings of the Association for Computational Linguistics:
  NAACL 2025}, 1370--1386.

\bibitem[{Kambhatla, Stewart, and Mihalcea(2022)}]{kambhatla2022surfacing}
Kambhatla, G.; Stewart, I.; and Mihalcea, R. 2022.
\newblock Surfacing racial stereotypes through identity portrayal.
\newblock In \emph{Proceedings of the 2022 ACM conference on fairness,
  accountability, and transparency}, 1604--1615.

\bibitem[{Kameswari and Mamidi(2021)}]{kameswari2021towards}
Kameswari, L.; and Mamidi, R. 2021.
\newblock Towards quantifying magnitude of political bias in news articles
  using a novel annotation schema.
\newblock In \emph{Proceedings of the International Conference on Recent
  Advances in Natural Language Processing (RANLP 2021)}, 671--678.

\bibitem[{Kameswari, Sravani, and Mamidi(2020)}]{kameswari2020enhancing}
Kameswari, L.; Sravani, D.; and Mamidi, R. 2020.
\newblock Enhancing bias detection in political news using pragmatic
  presupposition.
\newblock In \emph{Proceedings of the eighth international workshop on natural
  language processing for social media}, 1--6.

\bibitem[{Khan et~al.(2025)Khan, Sivakumar, Wang, Metcalf, Camacho, Theobald,
  Zappella, and Apostoloff}]{khan2025investigating}
Khan, F.~A.; Sivakumar, N.; Wang, Y.~O.; Metcalf, K.; Camacho, C.; Theobald,
  B.-J.; Zappella, L.; and Apostoloff, N. 2025.
\newblock Investigating Intersectional Bias in Large Language Models using
  Confidence Disparities in Coreference Resolution.
\newblock \emph{arXiv preprint arXiv:2508.07111}.

\bibitem[{Kuila et~al.(2024)Kuila, Jena, Sarkar, and
  Chakrabarti}]{kuila2024news}
Kuila, A.; Jena, S.; Sarkar, S.; and Chakrabarti, P.~P. 2024.
\newblock “News has Various Shades”: Quantifying and Analyzing Media Bias
  at Aspect-level Granularity.
\newblock \emph{ACM Journal on Computing and Sustainable Societies}, 2(4):
  1--51.

\bibitem[{Kunchukuttan(2020)}]{kunchukuttan2020indicnlp}
Kunchukuttan, A. 2020.
\newblock {The IndicNLP Library}.
\newblock
  \url{https://github.com/anoopkunchukuttan/indic_nlp_library/blob/master/docs/indicnlp.pdf}.

\bibitem[{Kuuse et~al.(2025)Kuuse, Kangur, Chakraborty, and
  Sharma}]{kuuse2025crowdsourced}
Kuuse, S.~F.; Kangur, U.; Chakraborty, R.; and Sharma, R. 2025.
\newblock Crowdsourced Fact-Checking or Biased Commentary? Analyzing Political
  Bias in Twitter's Community Notes.
\newblock In \emph{Companion Proceedings of the ACM on Web Conference 2025},
  2661--2669.

\bibitem[{Lee et~al.(2023)Lee, Kim, Kim, Kim, and Park}]{LEE2023128219}
Lee, S.; Kim, J.; Kim, D.; Kim, K.~J.; and Park, E. 2023.
\newblock Computational approaches to developing the implicit media bias
  dataset: Assessing political orientations of nonpolitical news articles.
\newblock \emph{Applied Mathematics and Computation}, 458: 128219.

\bibitem[{Lei and Huang(2024)}]{lei2024sentence}
Lei, Y.; and Huang, R. 2024.
\newblock Sentence-level Media Bias Analysis with Event Relation Graph.
\newblock In \emph{Proceedings of the 2024 Conference of the North American
  Chapter of the Association for Computational Linguistics: Human Language
  Technologies (Volume 1: Long Papers)}, 5225--5238.

\bibitem[{Lia et~al.(2025)Lia, Dipta, Zehady, Islam, Chakraborty, and
  Al~Wasif}]{lia2025read}
Lia, N.~J.; Dipta, S.~R.; Zehady, A.~K.; Islam, N.; Chakraborty, M.; and
  Al~Wasif, A. 2025.
\newblock Read Between the Lines: A Benchmark for Uncovering Political Bias in
  Bangla News Articles.
\newblock In \emph{Proceedings of the Second Workshop on Bangla Language
  Processing (BLP-2025)}, 61--79.

\bibitem[{L{\"u}{\"u}si et~al.(2024)L{\"u}{\"u}si, Kangur, Chakraborty, and
  Sharma}]{luusi2024political}
L{\"u}{\"u}si, L.; Kangur, U.; Chakraborty, R.; and Sharma, R. 2024.
\newblock Political Stance Detection in Estonian News Media.
\newblock In \emph{Proceedings of the 9th International Workshop on
  Computational Linguistics for Uralic Languages}, 12--28.

\bibitem[{Maity et~al.(2024)Maity, Sharma, Dhar, Abhishek, Gupta, and
  Varma}]{maity2024multilingual}
Maity, A.; Sharma, A.; Dhar, R.; Abhishek, T.; Gupta, M.; and Varma, V. 2024.
\newblock Multilingual bias detection and mitigation for Indian languages.
\newblock In \emph{Proceedings of the 7th Workshop on Indian Language Data:
  Resources and Evaluation}, 24--29.

\bibitem[{Medidoddi et~al.(2022)Medidoddi, Bantupalli, Chakraborty, and
  Mukherjee}]{10.1007/978-3-031-19097-1_33}
Medidoddi, V.; Bantupalli, J.; Chakraborty, S.; and Mukherjee, A. 2022.
\newblock Decoding Demographic un-fairness from Indian Names.
\newblock In Hopfgartner, F.; Jaidka, K.; Mayr, P.; Jose, J.; and Breitsohl,
  J., eds., \emph{Social Informatics}, 472--489. Cham: Springer International
  Publishing.
\newblock ISBN 978-3-031-19097-1.

\bibitem[{Navon, Reingold, and Shen(2025)}]{navon2025fairness}
Navon, I. R.~L.; Reingold, O.; and Shen, J.~H. 2025.
\newblock Fairness with respect to Stereotype Predictors: Impossibilities and
  Best Practices.
\newblock \emph{Transactions on Machine Learning Research}.

\bibitem[{Noh et~al.(2025)Noh, Han, Jeon, and Hong}]{noh2025biassist}
Noh, Y.-G.; Han, M.; Jeon, J.; and Hong, J.-H. 2025.
\newblock BIASsist: Empowering News Readers via Bias Identification,
  Explanation, and Neutralization.
\newblock In \emph{Proceedings of the 2025 CHI Conference on Human Factors in
  Computing Systems}, 1--24.

\bibitem[{Pryzant et~al.(2020)Pryzant, Martinez, Dass, Kurohashi, Jurafsky, and
  Yang}]{pryzant2020automatically}
Pryzant, R.; Martinez, R.~D.; Dass, N.; Kurohashi, S.; Jurafsky, D.; and Yang,
  D. 2020.
\newblock Automatically neutralizing subjective bias in text.
\newblock In \emph{Proceedings of the aaai conference on artificial
  intelligence}, volume~34, 480--489.

\bibitem[{Puro et~al.(2026)Puro, Kezo, Prashanth, Anandhan, Deikus, and
  Vveinhardt}]{puro2026challenges}
Puro, V.; Kezo, R.; Prashanth, N. N.~A.; Anandhan, H.; Deikus, M.; and
  Vveinhardt, J. 2026.
\newblock Challenges of Multilingual Research: A Discourse in the Context of
  Nagaland, India.
\newblock \emph{Social Inclusion}, 14.

\bibitem[{Rajwal(2024)}]{rajwal2024lihisto}
Rajwal, S. 2024.
\newblock LiHiSTO: a comprehensive list of Hindi stopwords.
\newblock \emph{Multimedia Tools and Applications}, 83(17): 50047--50059.

\bibitem[{Rawat and Vadivu(2022)}]{rawat2022media}
Rawat, S.; and Vadivu, G. 2022.
\newblock Media bias detection using sentimental analysis and clustering
  algorithms.
\newblock In \emph{Proceedings of international conference on deep learning,
  computing and intelligence: ICDCI 2021}, 485--494. Springer.

\bibitem[{Raza et~al.(2024)Raza, Bamgbose, Chatrath, Ghuge, Sidyakin, and
  Muaad}]{raza2024unlocking}
Raza, S.; Bamgbose, O.; Chatrath, V.; Ghuge, S.; Sidyakin, Y.; and Muaad, A.
  Y.~M. 2024.
\newblock Unlocking bias detection: Leveraging transformer-based models for
  content analysis.
\newblock \emph{IEEE Transactions on Computational Social Systems}, 11(5):
  6422--6434.

\bibitem[{Raza, Rahman, and Ghuge(2024)}]{raza2024dataset}
Raza, S.; Rahman, M.; and Ghuge, S. 2024.
\newblock Dataset Annotation and Model Building for Identifying Biases in News
  Narratives.
\newblock In \emph{Text2Story@ ECIR}, 5--15.

\bibitem[{Rodrigo-Gin{\'e}s, Carrillo-de Albornoz, and
  Plaza(2024)}]{rodrigo2024systematic}
Rodrigo-Gin{\'e}s, F.-J.; Carrillo-de Albornoz, J.; and Plaza, L. 2024.
\newblock A systematic review on media bias detection: What is media bias, how
  it is expressed, and how to detect it.
\newblock \emph{Expert Systems with Applications}, 237: 121641.

\bibitem[{Sabane, Litake, and Chadha(2023)}]{sabane2023breaking}
Sabane, M.; Litake, O.; and Chadha, A. 2023.
\newblock Breaking language barriers: A question answering dataset for hindi
  and marathi.
\newblock \emph{arXiv preprint arXiv:2308.09862}.

\bibitem[{Sen et~al.(2022)Sen, Ghatak, Khanuja, Rekha, Gupta, Dhakate, Sharma,
  and Seth}]{sen2022analysis}
Sen, A.; Ghatak, D.; Khanuja, G.; Rekha, K.; Gupta, M.; Dhakate, S.; Sharma,
  K.; and Seth, A. 2022.
\newblock Analysis of media bias in policy discourse in india.
\newblock In \emph{Proceedings of the 5th ACM SIGCAS/SIGCHI Conference on
  Computing and Sustainable Societies}, 57--77.

\bibitem[{Spinde et~al.(2023)Spinde, Hinterreiter, Haak, Ruas, Giese, Meuschke,
  and Gipp}]{spinde2023media}
Spinde, T.; Hinterreiter, S.; Haak, F.; Ruas, T.; Giese, H.; Meuschke, N.; and
  Gipp, B. 2023.
\newblock The media bias taxonomy: A systematic literature review on the forms
  and automated detection of media bias.
\newblock \emph{arXiv preprint arXiv:2312.16148}.

\bibitem[{Spinde et~al.(2022)Spinde, Krieger, Ruas, Mitrovi{\'c},
  G{\"o}tz-Hahn, Aizawa, and Gipp}]{spinde2022exploiting}
Spinde, T.; Krieger, J.-D.; Ruas, T.; Mitrovi{\'c}, J.; G{\"o}tz-Hahn, F.;
  Aizawa, A.; and Gipp, B. 2022.
\newblock Exploiting transformer-based multitask learning for the detection of
  media bias in news articles.
\newblock In \emph{International conference on information}, 225--235.
  Springer.

\bibitem[{Spinde et~al.(2021{\natexlab{a}})Spinde, Plank, Krieger, Ruas, Gipp,
  and Aizawa}]{spinde-etal-2021-neural-media}
Spinde, T.; Plank, M.; Krieger, J.-D.; Ruas, T.; Gipp, B.; and Aizawa, A.
  2021{\natexlab{a}}.
\newblock Neural Media Bias Detection Using Distant Supervision With {BABE} -
  Bias Annotations By Experts.
\newblock In Moens, M.-F.; Huang, X.; Specia, L.; and Yih, S. W.-t., eds.,
  \emph{Findings of the Association for Computational Linguistics: EMNLP 2021},
  1166--1177. Punta Cana, Dominican Republic: Association for Computational
  Linguistics.

\bibitem[{Spinde et~al.(2021{\natexlab{b}})Spinde, Rudnitckaia, Sinha, Hamborg,
  Gipp, and Donnay}]{spinde2021mbicmediabias}
Spinde, T.; Rudnitckaia, L.; Sinha, K.; Hamborg, F.; Gipp, B.; and Donnay, K.
  2021{\natexlab{b}}.
\newblock MBIC -- A Media Bias Annotation Dataset Including Annotator
  Characteristics.
\newblock arXiv:2105.11910.

\bibitem[{Spinde et~al.(2025)Spinde, Wu, Gaissmaier, Demartini, Echizen, and
  Giese}]{spinde2025enhancing}
Spinde, T.; Wu, F.; Gaissmaier, W.; Demartini, G.; Echizen, I.; and Giese, H.
  2025.
\newblock Enhancing media literacy: The effectiveness of (Human) annotations
  and bias visualizations on bias detection.
\newblock \emph{Information Processing \& Management}, 62(6): 104244.

\bibitem[{Wang et~al.(2025)Wang, Haider, Tohidi, Gupta, Zhang, Callison-Burch,
  Rothschild, and Watts}]{wang2025media}
Wang, J.~S.; Haider, S.; Tohidi, A.; Gupta, A.; Zhang, Y.; Callison-Burch, C.;
  Rothschild, D.; and Watts, D.~J. 2025.
\newblock Media Bias Detector: Designing and Implementing a Tool for Real-Time
  Selection and Framing Bias Analysis in News Coverage.
\newblock In \emph{Proceedings of the 2025 CHI Conference on Human Factors in
  Computing Systems}, 1--27.

\bibitem[{Warrens(2015)}]{warrens2015five}
Warrens, M.~J. 2015.
\newblock Five ways to look at Cohen's kappa.
\newblock \emph{Journal of Psychology \& Psychotherapy}, 5.

\bibitem[{Wessel et~al.(2023)Wessel, Horych, Ruas, Aizawa, Gipp, and
  Spinde}]{10.1145/3539618.3591882}
Wessel, M.; Horych, T.; Ruas, T.; Aizawa, A.; Gipp, B.; and Spinde, T. 2023.
\newblock Introducing MBIB - The First Media Bias Identification Benchmark Task
  and Dataset Collection.
\newblock In \emph{Proceedings of the 46th International ACM SIGIR Conference
  on Research and Development in Information Retrieval}, SIGIR '23,
  2765–2774. New York, NY, USA: Association for Computing Machinery.
\newblock ISBN 9781450394086.

\bibitem[{Wilkinson et~al.(2016)Wilkinson, Dumontier, Aalbersberg, Appleton,
  Axton, Baak, Blomberg, Boiten, da~Silva~Santos, Bourne
  et~al.}]{wilkinson2016fair}
Wilkinson, M.~D.; Dumontier, M.; Aalbersberg, I.~J.; Appleton, G.; Axton, M.;
  Baak, A.; Blomberg, N.; Boiten, J.-W.; da~Silva~Santos, L.~B.; Bourne, P.~E.;
  et~al. 2016.
\newblock The FAIR Guiding Principles for scientific data management and
  stewardship.
\newblock \emph{Scientific data}, 3(1): 1--9.

\bibitem[{Zhao et~al.(2018)Zhao, Wang, Yatskar, Ordonez, and
  Chang}]{zhao2018gender}
Zhao, J.; Wang, T.; Yatskar, M.; Ordonez, V.; and Chang, K.-W. 2018.
\newblock Gender Bias in Coreference Resolution: Evaluation and Debiasing
  Methods.
\newblock In \emph{Proceedings of the 2018 Conference of the North American
  Chapter of the Association for Computational Linguistics: Human Language
  Technologies, Volume 2 (Short Papers)}, 15--20.

\bibitem[{Zhou et~al.(2025)Zhou, Neumann, Garimella, and
  Gionis}]{zhou2025calibrated}
Zhou, T.; Neumann, S.; Garimella, K.; and Gionis, A. 2025.
\newblock Calibrated and Diverse News Coverage.
\newblock In \emph{Proceedings of the 34th ACM International Conference on
  Information and Knowledge Management}, 4509--4518.

\end{thebibliography}

\end{document}